\documentclass[final,5p,times,twocolumn,authoryear]{elsarticle}

\usepackage{amssymb}
\usepackage{amsmath}
\usepackage{xcolor}
\usepackage{subcaption}
\usepackage{threeparttable}
\usepackage{booktabs}

\usepackage[labelfont=bf, labelsep=period]{caption}
\usepackage{multirow}
\usepackage{hyperref}
\usepackage{url}
\usepackage{graphicx}
\usepackage{multirow}
\usepackage{enumitem}
\usepackage{float}

\journal{Neural Networks}

\begin{document}
\begin{frontmatter}

\title{Event-based Photometric Stereo via Rotating Illumination and Per-Pixel Learning}


\author[a]{Hyunwoo Kim}
\author[a]{Won-Hoe Kim}
\author[b]{Sanghoon Lee}
\author[c]{Jianfei Cai}
\author[d]{Giljoo Nam\corref{cor}}
\ead{namgiljoo@gmail.com}
\author[a]{Jae-Sang Hyun\corref{cor}}
\ead{hyun.jaesang@yonsei.ac.kr}

\cortext[cor]{Corresponding author\corref{cor}}




\affiliation[a]{organization={Department of Mechanical Engineering, Yonsei University},
            city={Seoul},
            postcode={03722}, 
            country={South Korea}}

            \affiliation[b]{organization={Department of Electrical and Electronic Engineering, Yonsei University},
            city={Seoul},
            postcode={03722}, 
            country={South Korea}}

            \affiliation[c]{organization={Department of Data Science and AI, Monash University},
            city={Melbourne},
            postcode={VIC 3800}, 
            country={Australia}}

\affiliation[d]{organization={Meta Reality Labs},
            country={United States}}

\begin{abstract}
Photometric stereo is a technique for estimating surface normals using images captured under varying illumination.
However, conventional frame-based photometric stereo methods are limited in real-world applications due to their reliance on controlled lighting, and susceptibility to ambient illumination.
To address these limitations, we propose an event-based photometric stereo system that leverages an event camera, which is effective in scenarios with continuously varying scene radiance and high dynamic range conditions.
Our setup employs a single light source moving along a predefined circular trajectory, eliminating the need for multiple synchronized light sources and enabling a more compact and scalable design.
We further introduce a lightweight per-pixel multi-layer neural network that directly predicts surface normals from event signals generated by intensity changes as the light source rotates, without system calibration.
Experimental results on benchmark datasets and real-world data collected with our data acquisition system demonstrate the effectiveness of our method, achieving a 7.12\% reduction in mean angular error compared to existing event-based photometric stereo methods.
In addition, our method demonstrates robustness in regions with sparse event activity, strong ambient illumination, and scenes affected by specularities.
\end{abstract}



\end{frontmatter}

\section{Introduction}\label{sec:introduction}

Photometric stereo \citep{woodham1980photometric} is a classical computer vision technique that enables the estimation of surface normals by capturing images under varying illumination directions.
Since surface normals are estimated at the same resolution as the input images on a per-pixel basis, photometric stereo enables the recovery of highly detailed, high-resolution surface information.
In particular, surface normal maps are essential for downstream tasks such as relighting, shape-from-shading, and depth reconstruction.

Conventional photometric stereo methods typically employ frame-based cameras and multiple light sources whose positions and directions are fixed.
However, this setup has several limitations when applied to real-world photometric stereo.
First, frame-based cameras are constrained by their limited dynamic range, which can lead to pixel saturation in the presence of strong ambient light.
Since the saturated pixels can significantly degrade the accuracy of surface normal estimation, the ideal photometric stereo setup requires a darkroom environment with controlled lighting conditions.
Second, conventional photometric stereo setups require multiple light sources, which necessitate careful calibration and controlled lighting conditions to ensure accurate illumination directions and intensities.
This requirement increases system complexity and limits the flexibility of deployment in unconstrained environments.
These factors hinder the practical deployment of frame-based photometric stereo in real-world scenarios.

\begin{figure*}[tbp]
    \centering
    \includegraphics[width=\linewidth]{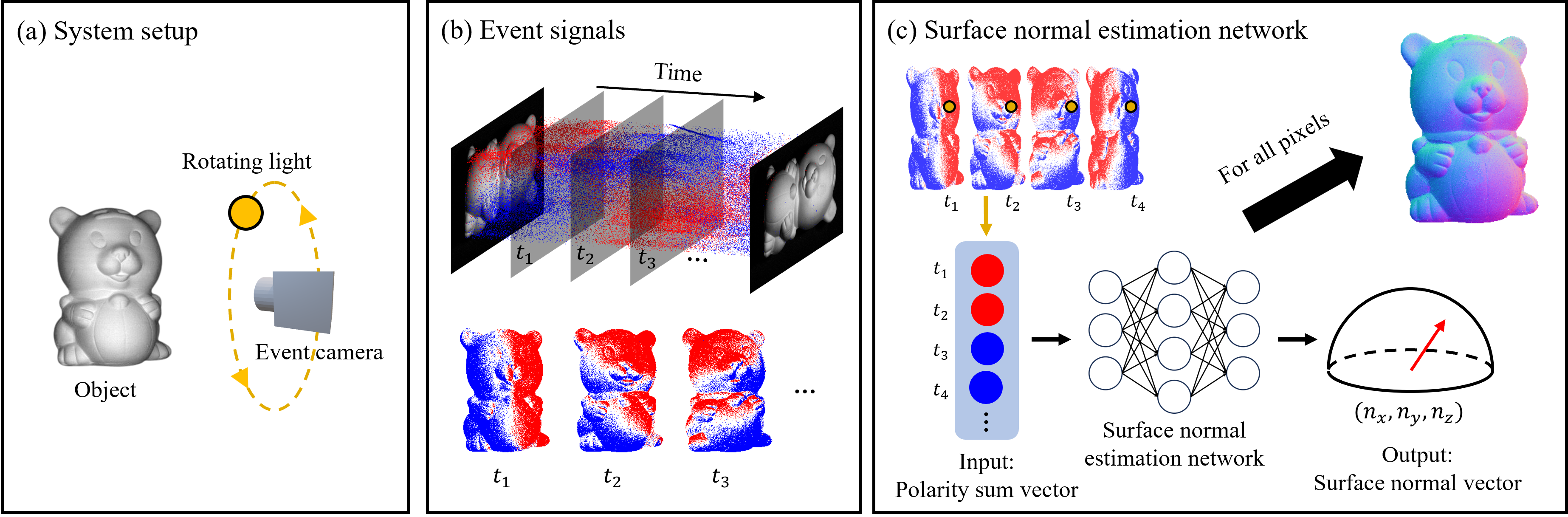}
    \caption{Overview of the proposed method. (a) Our system setup. A light source rotates around the optical axis of a fixed event camera. (b) Event signals observed over one rotation cycle of the light source. The signals are converted to a polarity-based event representation that encodes temporal illumination variations. (c) The polarity-based representation is aggregated at each pixel to form a polarity sum vector, which is used as the input to a surface normal estimation network. The network learns the nonlinear mapping from temporal event patterns to surface normals on a per-pixel basis and predicts dense surface normal maps without system calibration.}
    \label{fig:overview}
\end{figure*}

To overcome these limitations, recent studies \citep{ryoo2023event, yu2024eventps} have explored the use of event cameras, which offer higher dynamic range and temporal resolution than conventional frame-based cameras.
Event cameras typically offer a dynamic range exceeding 120 dB, while frame-based cameras are limited to around 60 dB \citep{gallego2020event}.
This high dynamic range enables event cameras to capture accurate information even in the presence of strong ambient light, where conventional sensors would suffer from saturation.
Moreover, unlike frame-based cameras that record absolute intensity at fixed frame rates, event cameras continuously monitor intensity and asynchronously output per-pixel events when the logarithmic intensity change exceeds a predefined contrast threshold.
Such continuous sensing capability is particularly effective in scenarios where scene radiance changes continuously over time.
In the context of photometric stereo, this temporal continuity enables a single moving light source to effectively emulate the effect of multiple fixed light sources used in traditional frame-based setups, thereby simplifying the hardware configuration while maintaining sufficient illumination diversity for accurate surface normal estimation.



Based on these advantages, we propose an event-based photometric stereo system that estimates surface normals from asynchronous event streams generated by a light source which rotates around the optical axis of the event camera.
Inspired by prior frame-based photometric stereo studies that demonstrated improved accuracy when multiple light sources were arranged along a circular trajectory to ensure diverse illumination directions \citep{zhou2010ring, fan2022near}, we adopt a similar principle by continuously rotating a single light source along a predefined circular path.
This design effectively emulates dense directional lighting while requiring only a single controllable light source, thereby eliminating the need for complex multi-light setups and synchronization.
Building upon this system, we reformulate the conventional frame-based photometric stereo formulation in the event domain to accommodate asynchronous event streams.
Through this reformulation, we demonstrate that surface normals can be directly estimated from event signals alone, without relying on absolute intensity measurements.

We further introduce a lightweight per-pixel multi-layer perceptron (MLP) model that directly predicts surface normals from the temporal patterns of event signals.
The MLP architecture is particularly well-suited for our formulation, as it can effectively learn the complex nonlinear relationship between the temporal patterns of event signals and surface normals.
To effectively capture the temporal dynamics of event streams induced by the rotating light source, we adopt a polarity-based event representation that encodes the time and polarity of events at each pixel over a rotation cycle.
Since this representation relies only on event polarity information, our method does not require explicit calibration of light directions and event contrast threshold.
Fig.~\ref{fig:overview} shows an overview of the proposed method.
To validate the effectiveness of our approach, we conduct comprehensive experiments on both benchmark datasets and real-world data acquired using our custom event-based photometric stereo setup.
The results demonstrate that our method achieves robust and accurate surface normal estimation under challenging conditions, including specularities, sparse event activities, and high illumination conditions.
In this paper, our contributions are summarized as follows:
\begin{itemize}
\item We present an event-based photometric stereo system that estimates surface normals from asynchronous event streams induced by a continuously rotating light source and reformulate the frame-based photometric stereo formulation in the event domain.
\item We introduce a lightweight per-pixel MLP network that directly predicts surface normals from event signals without system calibration.
\item We validate the proposed approach through comprehensive experiments on both benchmark and real-world datasets acquired with our custom event-based photometric stereo setup, demonstrating accurate and robust surface normal estimation under high dynamic range conditions.
\end{itemize}

\section{Related Works}\label{sec:related_works}

\subsection{Photometric stereo}
Photometric stereo estimates the surface normal at every pixel of an image by capturing multiple images of a static object under varying lighting directions from a fixed viewpoint. Some early methods adopt the Lambertian surface assumption to simplify the estimation problem \citep{woodham1980photometric, silver1980determining}. Other approaches attempt to model more complex reflectance behaviors using analytical formulations of the bidirectional reflectance distribution function (BRDF) \citep{tozza2016direct, chung2008efficient, ruiters2009heightfield}. Although these methods make the problem more tractable, they often lead to significant errors when applied to real-world surfaces, which commonly exhibit non-Lambertian characteristics such as specular highlights, shadows, and subsurface scattering. Moreover, although nonparametric BRDFs offer a more expressive representation of surface reflectance, they are challenging to employ in practice due to their high dimensionality and the difficulty of acquiring sufficient observations in uncontrolled environments.

To address the challenges posed by non-Lambertian reflectance, traditional photometric stereo methods have explored both physically based and statistical approaches. Physically based methods incorporate more sophisticated reflectance models to account for complex surface properties, such as rough surfaces with varying directional reflectance characteristics \citep{oren1995generalization}, spatially varying and arbitrary BRDFs \citep{goldman2009shape, hertzmann2003shape}, and bivariate BRDF approximations \citep{alldrin2008photometric}. In parallel, statistical approaches such as the hierarchical Bayesian model proposed by Ikehata et al. \citep{ikehata2012robust} demonstrated competitive performance even on surfaces exhibiting strong non-Lambertian behavior.

Despite these advances, traditional methods often require a large number of input images and struggle to handle high-frequency reflectance components. To mitigate these limitations, a follow-up method \citep{ikehata2014photometric} introduced a piecewise linear diffuse model, offering improved performance with fewer input images. However, existing approaches still face difficulties in addressing global illumination effects and in robustly estimating surface normals under complex lighting conditions. 

Learning-based approaches have emerged as powerful alternatives for addressing the limitations of traditional photometric stereo methods. One of the earliest deep learning methods in this domain is DPSN \citep{santo2017deep}, which introduced a deep neural network to learn a flexible mapping from complex reflectance observations to surface normals. Despite its effectiveness, DPSN is constrained by the requirement that the light directions must be fixed during training and testing, limiting its generalization capability. To overcome this constraint, PS-FCN \citep{chen2018ps} was introduced as a convolutional neural network (CNN) that takes as input a set of images captured under various lighting directions, along with their corresponding light direction vectors. The network aggregates these inputs in an order-agnostic manner and predicts surface normals globally, without relying on a fixed lighting setup. In contrast to PS-FCN, which predicts surface normals in a global manner, CNN-PS \citep{ikehata2018cnn} proposed a per-pixel estimation strategy based on a novel intermediate representation called the observation map. This representation encodes light directions and corresponding intensities at each pixel in a fixed-shape structure, enabling the network to handle an arbitrary number of input images in an order-agnostic manner. By feeding this per-pixel observation map into a CNN, the method effectively estimates surface normals while preserving local reflectance characteristics. Based on CNN-PS, PX-Net \citep{logothetis2021px} was proposed as an improved per-pixel observation map generation strategy for modeling global illumination effects. Unlike CNN-PS, which constructs observation maps from grayscale images, PX-Net generates the observation map from each RGB channel of the input images. This design removes the need for computationally expensive full-image rendering while still allowing the network to learn complex illumination interactions across the scene.

Recent studies have explored integrating photometric cues derived from multi-view photometric stereo with neural scene representations to improve surface reconstruction quality. In particular, multi-view approaches based on neural rendering frameworks such as 3D Gaussian splatting \citep{kerbl20233d} incorporate photometric regularization across multiple viewpoints to better constrain surface geometry \citep{ju2025photometric, kim2025multiview, deng2025depth}. By enforcing photometric consistency among multi-view observations, these methods improve geometric reconstruction accuracy. These studies highlight the importance of photometric cues within neural scene representation frameworks for accurate surface reconstruction.


\subsection{Event camera}
Unlike conventional cameras that output dense image frames, event cameras produce asynchronous streams of sparse events encoding per-pixel intensity changes. Each event contains the timestamp, pixel location, and polarity of the brightness change. Due to this fundamentally different data format, event streams are not directly compatible with standard computer vision pipelines or deep learning models, which typically require grid-structured inputs.

To address this, various event representations have been proposed to convert event data into formats suitable for learning-based methods. A common strategy is to aggregate events into image-like structures, making them compatible with existing architectures for frame-based vision tasks. Image-based representations include stacking events by polarity \citep{maqueda2018event}, by timestamp intervals \citep{wang2019event, deng2020amae, bai2022accurate}, or by event counts \citep{hu2020learning, messikommer2020event}. These methods enable spatial integration of events over a short time window. To better preserve temporal dynamics, other representations go beyond simple stacking. Voxel-based representations \citep{zihao2018unsupervised, ye2020unsupervised, rebecq2019events, baldwin2022time} partition the spatiotemporal event stream into a 3D volume, allowing the network to exploit fine-grained motion and intensity changes. Time surface-based approaches \citep{benosman2013event, lagorce2016hots, alzugaray2018ace, afshar2019investigation} encode the recency of events by storing the timestamp of the last event at each pixel, producing a smooth surface that evolves over time. These representations serve as the foundation for applying event data to image- or video-based deep learning models, enabling learning-based methods to leverage the high temporal resolution and dynamic range of event cameras in a structured form.

With the development of event representation techniques, event cameras have been increasingly applied to a wide range of computer vision tasks. Recent works have demonstrated their effectiveness in image reconstruction \citep{rebecq2019high, ercan2024hypere2vid}, feature tracking \citep{gehrig2020eklt, messikommer2023data}, depth estimation \citep{muglikar2021event, ghosh2022multi}, and 3D reconstruction \citep{muglikar2023event, klenk2023nerf}. In particular, photometric stereo has emerged as a promising application for event cameras due to their robustness under challenging lighting conditions and their ability to capture rapid brightness changes with high temporal resolution. This enables precise surface normal estimation even in dynamic or high-contrast environments, where conventional cameras often suffer from motion blur or saturation. EFPS-Net \citep{ryoo2023event} combines observations from an event camera and an RGB camera that share the same viewpoint. The high dynamic range, sparse observation map from the event camera is fused with the dense, low dynamic range observation map from the RGB camera to produce accurate surface normal predictions. Another method, EventPS \citep{yu2024eventps}, leverages a null-space vector formulation to propose both optimization-based and learning-based strategies for surface normal estimation from event data. These studies highlight the growing interest in event-based photometric stereo and its potential to overcome the limitations of traditional RGB-based approaches in dynamic and high-contrast lighting conditions. 
\section{Methods}\label{sec:methods}

\subsection{Problem Formulation}\label{sec:problem_formulation}

\paragraph{Image Formation Model}

Classic photometric stereo \citep{woodham1980photometric, silver1980determining} aims to reconstruct surface normals of an object with a Lambertian reflectance by utilizing images captured under varying lighting directions. Assuming a Lambertian surface, the observed intensity is linearly dependent on the dot product between the surface normal and the light direction, modulated by the object's BRDF. This relationship is commonly expressed as
\begin{equation}
    \label{equation:photometric_stereo_1}
        I_j = A_j \cdot \rho(n, l_j, v) \cdot max(n^T \cdot l_j, 0),
\end{equation}
where $I_j$ denotes the observed intensity under the $j$-th light source.
$A_j$, and $l_j$ denote the absolute intensity and the direction of the light source, respectively. $v$ is the viewing direction of the camera,  $\rho$ is a BRDF, and $n$ is a unit surface normal vector. The $max(n^T\cdot l_j, 0)$ term accounts for the effects of shading and attached shadow. Assume that all light sources have equal intensity, the BRDF is isotropic, and there is no attached shadow, Eq.~\ref{equation:photometric_stereo_1} is simplified as follows.

\begin{equation}
    \label{equation:phometric_stereo_2}
        I_j = \rho \space n^T \cdot l_j.
\end{equation}
The surface intensity $I_j$ can be directly measured from the captured images, and light direction $l_j$ can be obtained via light source calibration \citep{yang2023point}. Thus, surface normal $n$ and BRDF $\rho$ can be obtained by inversely solving Eq .\ref{equation:phometric_stereo_2}  with measurements captured under varying illumination.

\paragraph{Event Camera}

Unlike conventional RGB cameras, event cameras asynchronously detect changes in logarithmic intensity, $L=log(I)$ at each pixel. An event signal $e_k = (t_k, \mathbf{x_k}, p_k)$ is triggered at the pixel $\mathbf{x_k}=(x_k, y_k)^T$ and timestamp $t_k$ with polarity $p_k$ which indicates whether the radiance has decreased $(-1)$ or increased $(+1)$. This occurs when the change in logarithmic radiance at the same pixel reaches a temporal contrast threshold $C$, i.e.,

\begin{equation}
    \label{equation:event_cam_1}
        \Delta L(\mathbf{x_k}, t_k) = L(\mathbf{x_k}, t_k) - L(\mathbf{x_k}, t_{k-1}) = p_k C.
\end{equation}
Assuming a noise-free scenario, the event data accumulated from the initial timestamp $t_0$ can be used to reconstruct the intensity at each timestamp based on the initial intensity information. The intensity at any given timestamp  can be recovered as follows

\begin{equation}
    \label{equation:event_cam_2}
    \begin{split}
                L(t_N)& = \sum_{k=1}^{N}p_k C + L(t_0), \\
I(t_N) = exp(L(t_N))&=I(t_0)exp(\sum_{k=1}^{N}p_k C) = I(t_o) E_k,
    \end{split}
\end{equation}
where $E_k = exp(sum_{k=1}^{N}p_k C)$ denotes the accumulated exponential contrast induce by a stream of events up to the event timestamp $t_k$.

\paragraph{Intensity-Normal Formulation with pre-defined light trajectory} 


\begin{figure}[t]
    \centering
    \includegraphics[width=\linewidth]{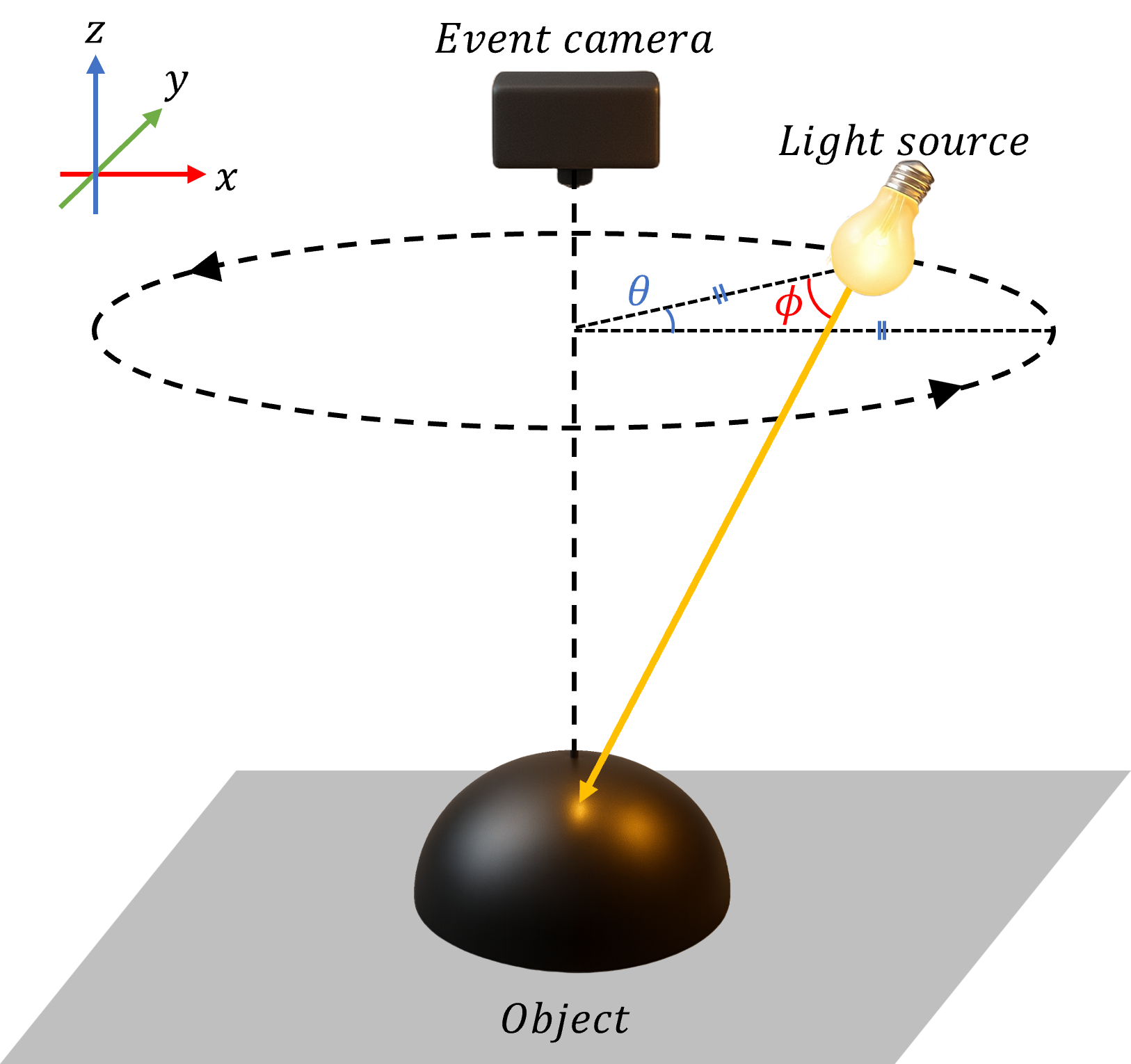}
    \caption{Schematic diagram of our system. The system consists of a fixed event camera and a light source. The light source rotates around the optical axis of the event camera at a constant angular speed.}
    \label{fig:system_schematic}
\end{figure}

In this paragraph, we describe the hardware configuration of our acquisition system and derive an analytic solution to estimate surface normals from the signals captured by our system. We constructed a system in which the light source rotates around the optical axis of the camera. Assuming that the light source rotates at a constant angular velocity, its direction can be parameterized based on the system configuration, as illustrated in Fig.~\ref{fig:system_schematic}. The elevation angle, $\phi$ is fixed, while the azimuth angle $\theta$ varies linearly with time.  For convenience, let $t$ be the normalized timestamp, then the light direction $l(t)$ is defined as follows

\begin{equation}
    \label{equation:light_direction}
    l(t) = 
    \begin{bmatrix}
        \cos\phi(t)\cos\theta(t) \\ 
        \cos\phi(t)\sin\theta(t) \\
        \sin\phi(t)
    \end{bmatrix}
    =     \begin{bmatrix}
        \cos \phi \cos t \\
        \cos \phi \sin t \\ 
        \sin \phi
    \end{bmatrix}.
\end{equation}
Under this parameterization, the light direction $l(t)$ traces a circular trajectory around the optical axis while maintaining a constant elevation. Applying the trajectory used in our scenario to the intensity formulation yields the following expression
\begin{equation}
    \label{equation:intensity_normal}
    \begin{split}
        I(t)  &=  \rho(n_x \cos \phi \cos t + n_y \cos\phi \sin t + n_z \sin \phi) \\ &= \rho\cos\phi \sqrt{\beta}\cos(t - \alpha ) + \rho\sin\phi \sqrt{1- \beta},
    \end{split}
\end{equation}
where $\alpha$ and $\beta$ represents $tan^{-1} (n_y / n_x)$ and $n_x ^ 2+n_y^2$, respectively. This formulation establishes an explicit relationship between the observed intensity and the surface normal, enabling per-pixel normal estimation from temporal intensity variations under a known lighting trajectory.

\paragraph{Event-Normal Formulation}
Conventional cameras integrate incoming light over the duration of the exposure—from the moment the shutter opens to when it closes. In our setup, where the light source continuously rotates, this integration process leads to scene variations during a single exposure.
As a result, the captured image represents an average of intensities over a range of light source positions and does not correspond to a specific light direction at a precise time.

In contrast, event cameras continuously respond to changes in intensity, enabling the capture of temporally precise information at specific light source positions—even during rotation. However, event cameras do not measure absolute intensity values, but rather capture the relative changes in intensity between consecutive events. Consequently, conventional intensity-normal formulations cannot be directly applied for normal estimation.

Despite this limitation, surface normals can be inferred from event data using a formulation analogous to that of intensity-based photometric stereo. As shown in Eq.~\ref{equation:event_cam_2}, the exponential of the accumulated event polarities represents the relative change in intensity with respect to the initial intensity. By incorporating this relationship into Eq.~\ref{equation:intensity_normal}, the exponential of the event polarity sum can be reformulated as follows.

\begin{equation}
    \label{equation:event_normal1}
    \begin{split}
        E(t_k) &= \frac{I(t_k)}{I(t_0)} \\ 
        &= \frac{\cos\phi \sqrt{\beta}\cos(t_k - \alpha ) + \sin\phi \sqrt{1- \beta}}{\cos\phi \sqrt{\beta}\cos(t_0 - \alpha ) + \sin\phi \sqrt{1- \beta}}.
    \end{split}
\end{equation}
Similar to the intensity-normal formulation, $E(t_k)$ can be expressed as a cosine function of time.
\begin{equation}
\label{equation:event_normal2}
    \begin{array}{c}
        E(t) = E_{amp}\cos(t - E_{\phi}) + E_0, \\[1.5ex]
        E_{amp} = \dfrac{\cos\phi \sqrt{\beta}}{\cos\phi \sqrt{\beta} \cos(t_0 - \alpha ) + \sin\phi \sqrt{1 - \beta}}, \\[2.5ex]
        E_{\phi} = \alpha, \\[1.5ex]
        E_{0} = \dfrac{\sin\phi \sqrt{1 - \beta}}{\cos\phi \sqrt{\beta} \cos(t_0 - \alpha) + \sin\phi \sqrt{1 - \beta}}.
    \end{array}
\end{equation}
Let $E_{amp}$, $E_{\phi}$, and $E_0$ denote amplitude, phase shift, and offset, respectively. 
Estimating these cosine parameters from the observed event stream, the surface normal can be analytically recovered as a function of these parameters and the constant elevation angle $\phi$. From Eq.~\ref{equation:event_normal2}, $\alpha$ and $\beta$ can be expressed in terms of these parameters.
\begin{equation}
\label{equation:alpha_beta}
    \begin{array}{c}
        \alpha = \tan^{-1} \dfrac{n_y}{n_x} = E_\phi, \\
        \beta = n_x^2 + n_y^2 =  \dfrac{(E_0 \tan\phi)^2}{(E_0 \tan\phi)^2 + {E_0}^2}.
    \end{array}
\end{equation}

Finally, the resulting surface normal vector can be analytically expressed as follows.
\begin{equation}
\label{equation:surface_normal}
    \begin{array}{c}
        n_x = \dfrac{E_{amp} \tan\phi \cos E_{\phi}}{\sqrt{(E_{amp} \tan\phi)^2 + E_{0}^2}}, \\[3.5ex]
        n_y = \dfrac{E_{amp} \tan\phi \sin E_{\phi}}{\sqrt{(E_{amp} \tan\phi)^2 + E_{0}^2}}, \\[3.5ex]
        n_z = \dfrac{E_{0}}{\sqrt{(E_{amp} \tan\phi)^2 + E_{0}^2}}.
    \end{array}
\end{equation}

\subsection{Event-normal network}\label{sec:event_normal_network}

In Section~\ref{sec:problem_formulation}, we demonstrated that surface normals can be estimated from event signals under the assumption of a Lambertian surface and the absence of global illumination effects such as cast shadows and inter-reflections. However, real-world surfaces rarely adhere to these ideal conditions; non-Lambertian phenomena including specular highlights, shadows, and inter-reflections are commonly present. Previous studies in photometric stereo with conventional cameras (e.g., CNN-PS \citep{ikehata2018cnn}, PS-FCN \citep{chen2018ps}) have shown that deep learning-based approaches exhibit strong robustness against such non-idealities. Inspired by these findings, we hypothesize that event signals, similar to RGB images, can also benefit from learning-based methods in handling complex lighting effects. Based on this observation, we propose a learning-based framework for surface normal estimation from event data.

The proposed method is a MLP network designed to predict surface normals from event signals collected over a single rotation cycle of a light source following a predefined trajectory. The network operates on a per-pixel basis and produces a normal map comprising surface normals estimated at each pixel location. Since surface normals can be inferred from the exponential of the accumulated event polarities, the network architecture is designed to effectively capture and model the polarity dynamics through deep feature representation.

The event signal used as input to the network is constructed based on a fixed contrast threshold and a predefined lighting trajectory, and is consistently applied during both the training and inference phases. During training, the network learns a mapping between the event signal and the corresponding surface normal under this fixed configuration. Importantly, the model implicitly encodes both the contrast threshold and the lighting behavior, enabling it to generalize within the constraints of the defined system setup.

The proposed network takes as input the event data observed at each pixel over a full rotation cycle of the light source and outputs the corresponding surface normal vector $n$. To enable accurate prediction, it is essential to construct an effective representation of the input event data. At the pixel level, the raw event stream consists of discrete events, each defined by a polarity and its corresponding timestamp. Thus, it is critical to encode these sparse and asynchronous signals into a meaningful representation that preserves both their spatiotemporal structure and polarity information.

To address this, we adopt a polarity-based event representation $P$, which aggregates the temporal dynamics of event polarities observed during the light source's motion. This representation is specifically designed to capture the relative intensity variations encoded in the event stream, thereby facilitating more effective learning of the underlying surface geometry.

\begin{figure}[t]
    \centering
    \begin{subfigure}[b]{0.45\linewidth}
        \includegraphics[width=\linewidth]{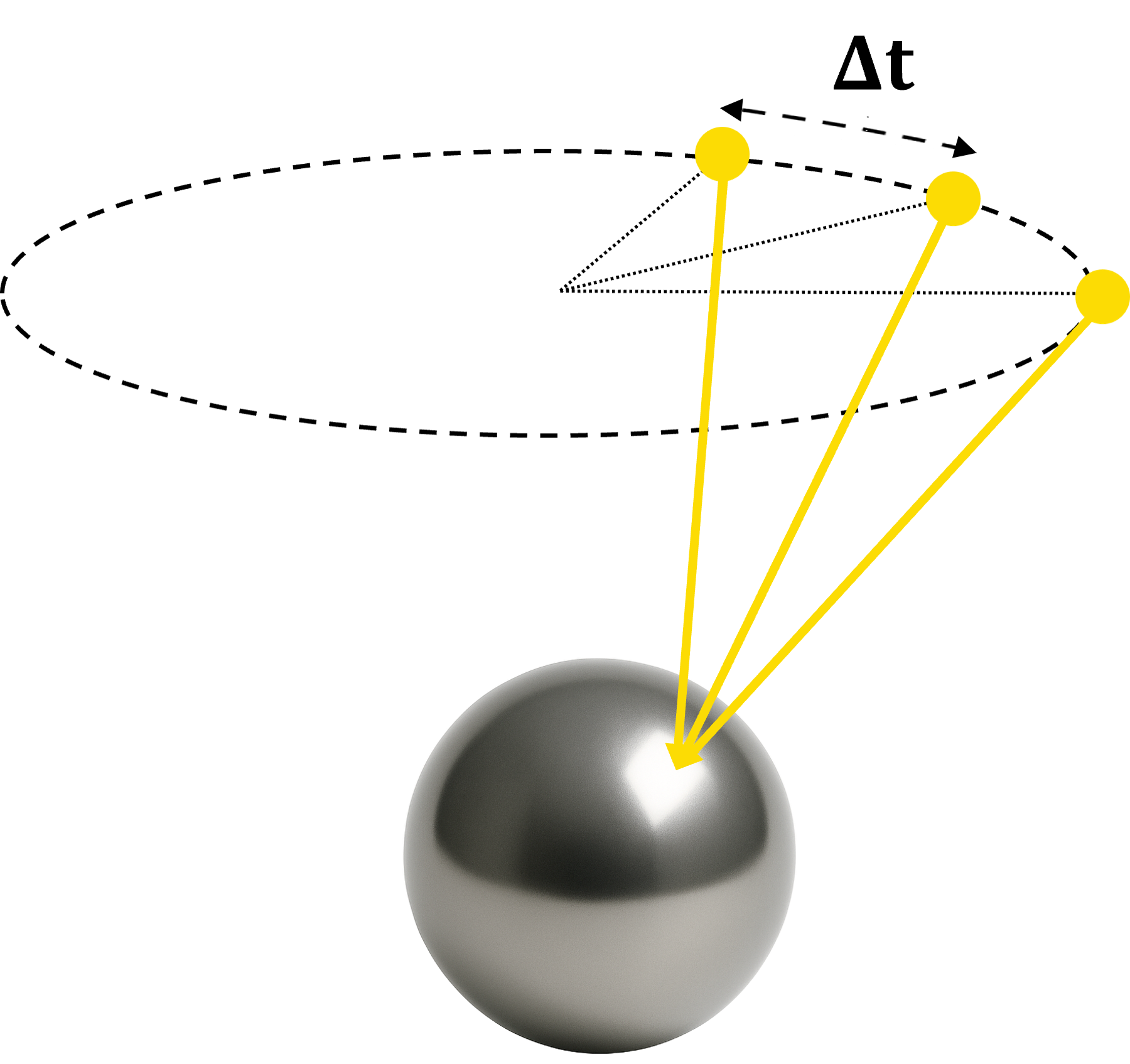}
        \caption{}
        \label{fig:left}
    \end{subfigure}
    \hfill
    \begin{subfigure}[b]{0.45\linewidth}
        \includegraphics[width=\linewidth]{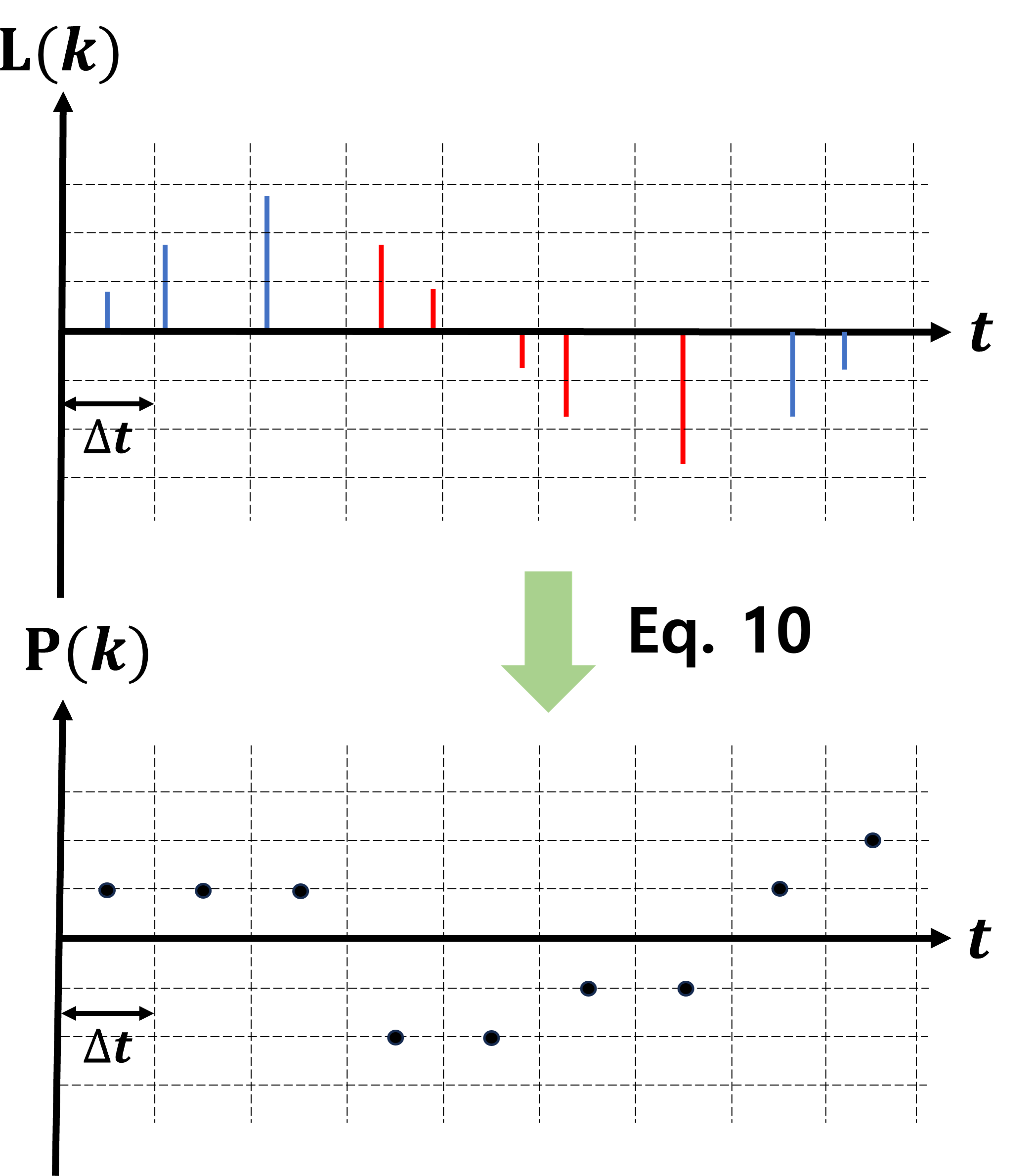}
        \caption{}
        \label{fig:right}
    \end{subfigure}
    \caption{Illustration of the process of generating event representations. (a) Intensity variations are induced by a rotating light source. (b) Events triggered by these variations are temporally sliced to compute the event representation.}
    \label{fig:event_representation}
\end{figure}

The event representation $P$ is defined as an event polarity sum vector of length $M$. To construct $P$, one full cycle of the light source's motion is temporally divided into $M$ uniform segments. For each time index $k$, the $k$-th element of $P$, denoted as $P(k)$, where $k=\{1,2,...,M\}$, represents the sum of event polarities within a segment. This can be mathematically expressed as
\begin{equation}
    \label{equation:accum_event}
    P(k) = \sum_{t=(k-1) \Delta T}^{k \Delta T} p(t).
\end{equation}
where $\Delta T$ denotes the temporal interval between consecutive time indices. Fig.~\ref{fig:event_representation}  illustrates the event polarity sum vector $P$ in our scenario, revealing how polarity information accumulates over time across one full cycle of the light source trajectory.
Since $P$ is determined only by event polarities, the method does not require explicit calibration of the light directions and the contrast threshold.
These factors are implicitly encoded by the fixed temporal segmentation over one rotation cycle and learned by the network, thereby removing the need for calibration while assuming only a repeatable rotation cycle for time indexing.

The architecture of the proposed network consists of a sequence of fully connected (dense) layers. To mitigate overfitting, dropout is applied after each layer, and Tanh activation functions are used for all intermediate outputs. The final layer produces a three-dimensional output corresponding to the x, y, and z components of the surface normal vector. To ensure that the output represents a unit normal vector, the predicted vector is normalized using the L2 norm. Additionally, since the surface normals captured by the camera are expected to face toward the camera, the z-component of the predicted normal is constrained to be positive by flipping the vector when necessary. The number of layers and the number of nodes in each layer were determined through experimental evaluation, and the final architecture is summarized in Table~\ref{tab:EMLP_structure}. The input dimension was set to 96, and the dropout rate was fixed at 0.2 throughout the network.

\begin{table}[t]
    \centering
    \caption{The network structure. The numbers in the parentheses represent the number of the output dimensions. Dropout rates for training are set to 0.2 for all Dense layers. The input dimension of the first Dense layer is set to 96.}
    \label{tab:EMLP_structure}
    \renewcommand{\arraystretch}{1.2}
    \resizebox{0.5\linewidth}{!}{%
    \begin{tabular}{cc}
    \toprule
    Layer & Description \\
    \midrule
    1 & Dense-(4096), Tanh, Dropout \\
    2 & Dense-(4096), Tanh, Dropout \\
    3 & Dense-(2048), Tanh, Dropout \\
    4 & Dense-(2048), Tanh, Dropout \\
    5 & Dense-(2048), Tanh, Dropout \\
    6 & Dense-(3) \\
    \bottomrule
    \end{tabular}%
    }
\end{table}

\begin{figure*}[htbp]
    \centering
    \includegraphics[width=\textwidth]{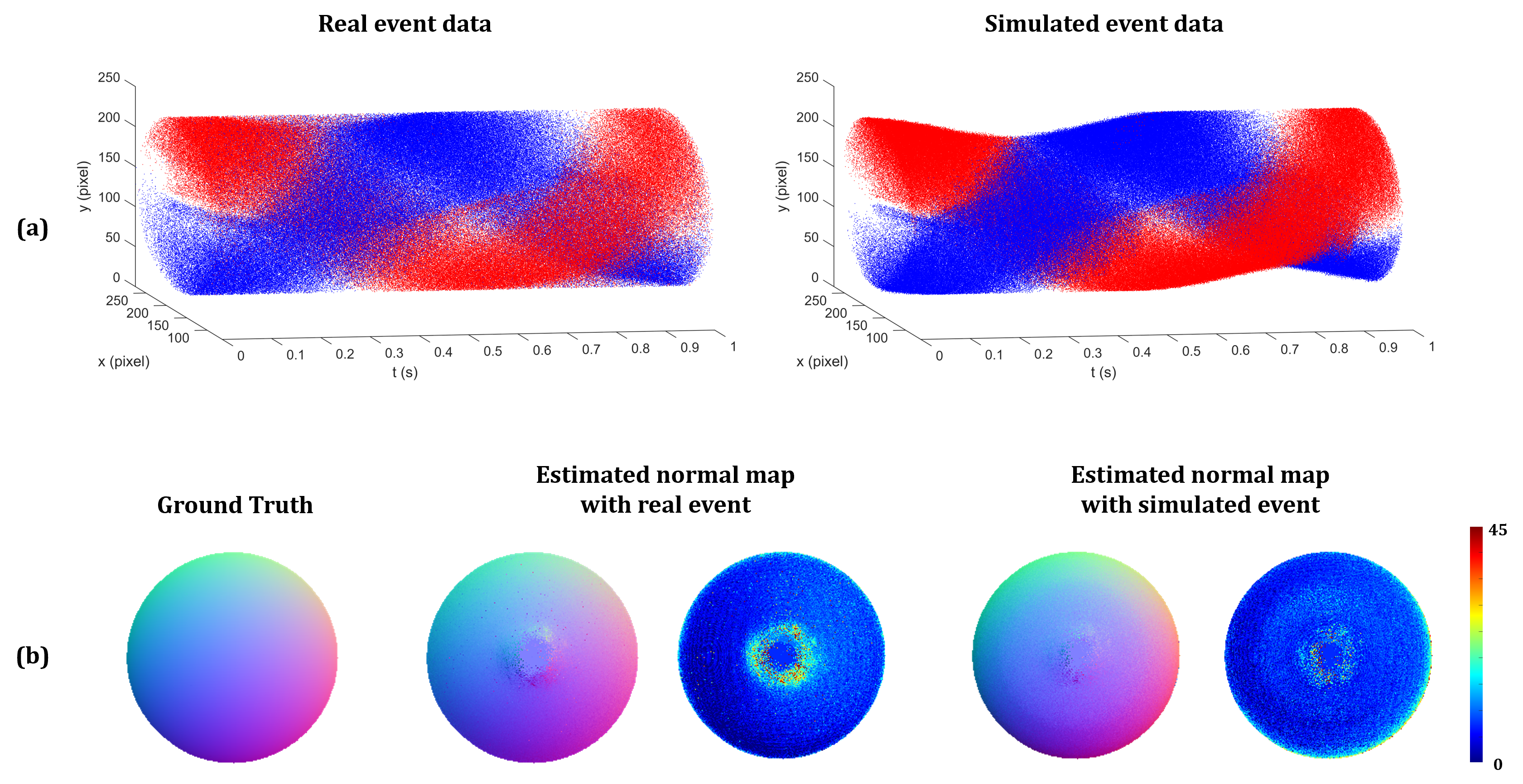}
    \caption{Comparison between simulated and real event data over one cycle of the light source rotation. (a) Event signals during one cycle. Blue points represent ON events, and red points represent OFF events. (b) Estimated surface normal maps and corresponding angular error maps for each event signal.}
    \label{fig:real_vs_sim}
\end{figure*}

\subsection{Loss function}
In photometric stereo, the surface normal vector is a unit vector, meaning that its directional orientation is more important than the magnitude of its individual components. Accordingly, we employed a cosine similarity loss to minimize the angular difference between the predicted surface normal $\hat{n}$ and the ground truth normal $n$. Since cosine similarity ranges from –1 to 1 and increases as the angle between the two vectors decreases, we used the cosine similarity term by subtracting the cosine similarity from 1. This formulation encourages alignment between the predicted and ground truth vectors and is mathematically defined as shown in Equation~\ref{equation:cosine_loss}.

\begin{equation}
    \label{equation:cosine_loss}
    \mathcal{L}_c = 1 - \frac{n \cdot \hat{n}}{\lVert n \rVert \lVert \hat{n} \rVert} = 1 - n \cdot \hat{n}.
\end{equation}

\section{Datasets}\label{sec:datasets}

To effectively train learning-based photometric stereo models, it is necessary to construct a large-scale dataset. Conventional photometric stereo datasets \citep{matusik2003data, johnson2011shape, shi2016benchmark} typically consist of RGB images captured under several discrete light sources with different directions, along with ground truth surface normal maps. However, our method estimates surface normals from event signals generated by the continuous motion of a light source, and therefore, these datasets cannot be used directly. To obtain the required event signals, we built a system that records the output of an event camera while varying the lighting direction over time.
There are two ways to obtain event data: recording real scenes with an event camera or generating synthetic event streams from time-sequential images using an event simulator. Since collecting a large-scale dataset of real event data with corresponding ground truth surface normals is costly expensive and time-consuming, we trained our model on a synthetic dataset and evaluated it using real event data collected with our system.

To obtain ground truth surface normal maps, we considered two approaches. The first method involves capturing the 3D geometry of an object using a precise 3D scanner and computing surface normals from the reconstructed mesh. The second method involves printing a 3D model of the object using a 3D printer and then computing surface normals from the camera’s viewpoint based on known geometric relationships, including the rotation matrix, translation vector, and camera intrinsics. In our work, we adopted the second approach and used Mitsuba 3 renderer \citep{jakob2022mitsuba3} to generate normal maps from the 3D models under the specified camera parameters.

\subsection{Train dataset}

The training dataset was constructed synthetically by simulating event data from time-sequential grayscale images rendered in Blender \citep{blender}. To reflect the high dynamic range (HDR) characteristics of event cameras and to prevent saturation at high exposure levels, we rendered all images using the HDR EXR format. Event signals were generated by interpolating the log intensity between consecutive frames and triggering an event whenever the log intensity change exceeded a contrast threshold, following the standard event camera operation model. Since the contrast threshold in real event cameras varies slightly over time rather than remaining constant, we simulated this behavior by sampling the contrast threshold from a Gaussian distribution with a given mean and standard deviation. To verify that the simulated events reflect the behavior of real event data, we compared them with real event data captured from the same viewpoint. The simulated events were generated using grayscale images rendered from the same view as the real event recordings. Fig.~\ref{fig:real_vs_sim} shows a comparison between the simulated and real event data collected during one cycle of the light source rotation, as well as the corresponding surface normal maps estimated from each event data. The normal maps were computed using the cosine regression method with an exponential of the event polarity sum, as described in Section~\ref{sec:methods}.

\begin{figure}[t]
    \centering
    \includegraphics[width=\linewidth]{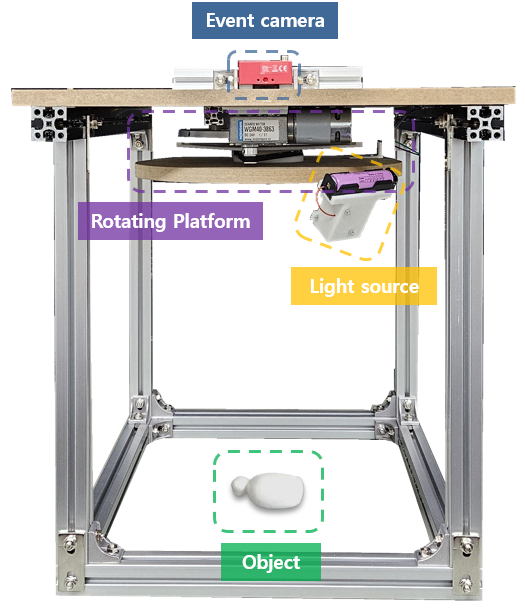}
    \caption{Our experiment setting. The rotating platform consists of a DC motor, a timing belt, and a hollow disc. The disc is connected to the motor through the timing belt, and rotates together. The light source is attached to the edge of the disc.}
    \label{fig:out_setup}
\end{figure}

However, there is a significant difference between the event simulator and real-world event cameras in terms of how the initial reset value for event triggering is defined at the start of recording. In real event data acquisition, the recording typically begins while the light source is already in motion, meaning that the initial reset value at each pixel reflects the log intensity state resulting from the previous cycle. In contrast, simulated event data initializes the reset value using the intensity of the first grayscale image, which does not reflect the temporal dynamics present in real-world operation. As a result, the events simulated during the first rotation cycle are likely to be inaccurate. To mitigate this issue and improve the reliability of the simulated events, we used the events generated during the second rotation cycle.

To construct the dataset, we used two categories of 3D models: Blobby objects from \citep{johnson2011shape} and Sculpture objects from \citep{wiles2017silnet}. The training dataset consists of 20 Blobby objects and 20 Sculpture objects. Each object was imported into Blender with a known pose, and the camera and light source were arranged according to our system. To ensure that the model could generalize to various surface reflectance properties, the BRDF for each object was randomly assigned. For each object, we captured 200 grayscale images during two cycles of the light source, and event data were then simulated from these frames. The ground truth surface normals were obtained by rendering the scene in Mitsuba 3 using the known pose of the object.

\subsection{Validation dataset}

\begin{figure}[t]
    \centering
    \includegraphics[width=\linewidth]{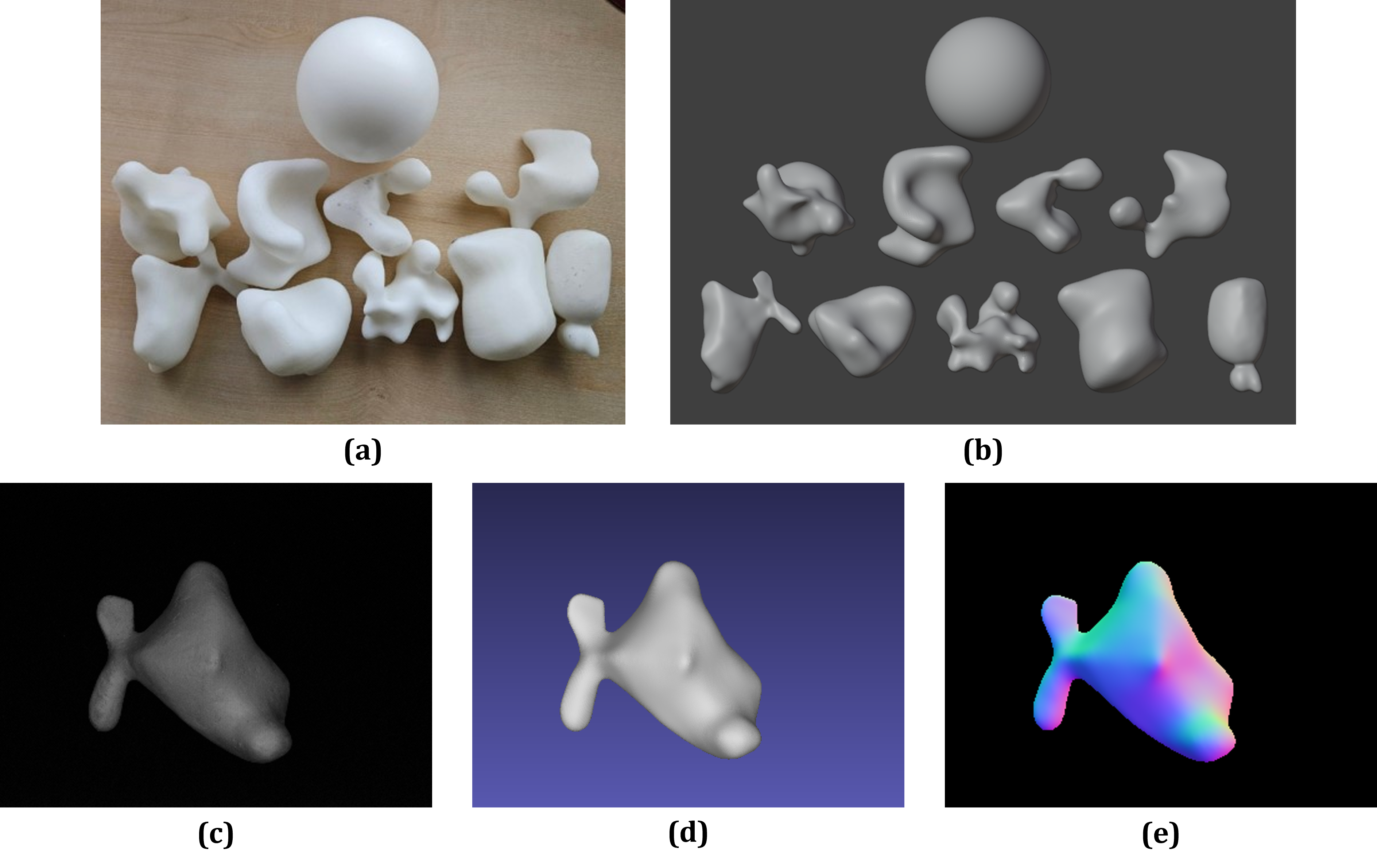}
    \caption{Dataset creation with Blobby obejcts. (a) 3D-printed objects in the real world. (b) Corresponding ground truth 3D models. (c) The image captured in the real world. Averaged during one cycle of the light source. (d) Rendered scene with the same pose as captured image. (e) Corresponding ground truth surface normal map.}
    \label{fig:3d_printed_objects}
\end{figure}

\begin{figure}[t]
    \centering
    \includegraphics[width=0.7\linewidth]{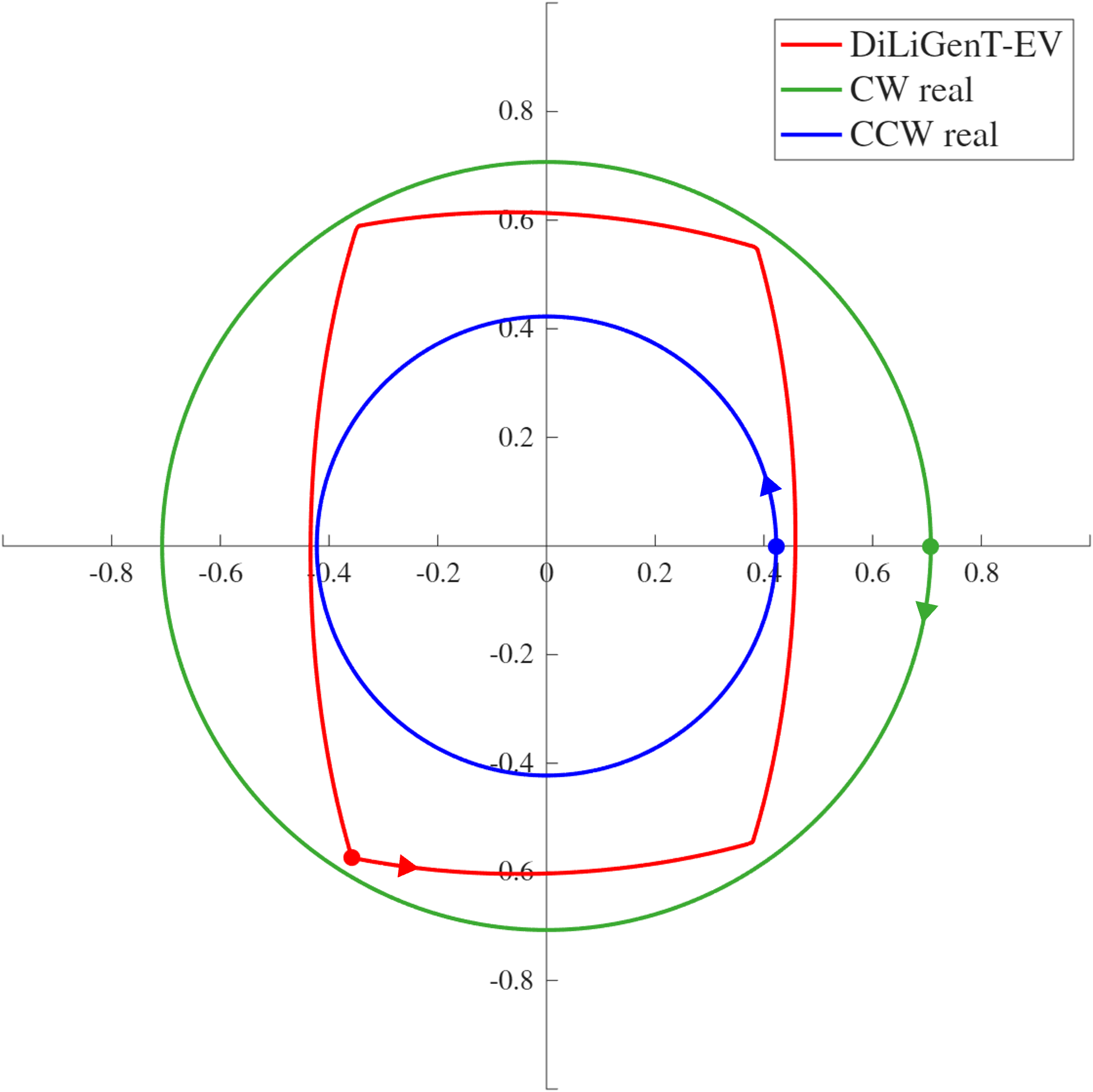}
    \caption{Light source trajectories used in each dataset. The bold circular markers indicate the initial point of each trajectory, and the arrows indicate the direction of the light source motion along each trajectory.}
    \label{fig:light_trajectory}
\end{figure}

To evaluate the real-world performance of the proposed method, we constructed a validation dataset using event data captured in the real world. For this purpose, we designed a custom validation platform, as illustrated in Fig.~\ref{fig:out_setup}. In our setup, an event camera captures the object from a top-down view, while a point light source rotates counterclockwise around the optical axis of the camera. The light source was mounted on the perimeter of a hollow disk. The disk was driven at a constant angular velocity by a DC motor connected through a timing belt. The rotation period of the light source is measured using a photointerrupter sensor. We used a DAVIS 346 mono event camera and a high-intensity LED point light source for the photometric stereo setup. For target objects, 10 Blobby objects from \citep{johnson2011shape} were 3D-printed. All objects were rescaled to have a maximum size of 100mm $\times$ 100mm $\times$ 100mm to fit within the field of view of our system. All 3D-printed objects are shown in Fig.~\ref{fig:3d_printed_objects}(a).

To obtain ground truth surface normal maps, it is necessary to estimate the relative transformation between the camera and object coordinate systems. Given an image of the target object, the rotation matrix and translation vector between the 3D model and the camera can be estimated by performing image-to-geometry registration based on mutual information \citep{corsini2009image}. Since the DAVIS 346 event camera features both an active pixel sensor (APS) and a dynamic vision sensor (DVS), it allows for simultaneous acquisition of event data and intensity images. Using the captured images, we estimated the rotation matrix and translation vector between the 3D model and the camera through image alignment tool in MeshLab \citep{cignoni2008meshlab}. Based on this information, we rendered the scene in Mitsuba 3 \citep{jakob2022mitsuba3}, and generate ground truth surface normal maps. Fig.~\ref{fig:3d_printed_objects}(d)\&(e) show an example of a rendered scene and a ground truth surface normal map.

We selected the EventPS dataset \citep{yu2024eventps} as the baseline for comparison, and used the same validation datasets as described in their work: the DiLiGenT-EV semi-real dataset and a real-world event dataset. The DiLiGenT-EV dataset was constructed by selecting images corresponding to the outermost lighting directions from the DiLiGenT dataset \citep{shi2016benchmark}, and converting them into event streams using an event simulator \citep{rebecq2018esim}. In the real dataset, event data were collected using a system in which a point light source rotates clockwise around the optical axis of a fixed event camera. This system is conceptually similar to our own real-world system, except for the rotation direction and elevation angle of the light source. Fig.~\ref{fig:light_trajectory} shows the light source trajectories used in each dataset. We used the following three datasets as our validation datasets: 1) DiLiGenT-EV dataset, 2) CW real dataset, and 3) CCW real dataset.

\begin{table*}[t]
    \centering
    \caption{Quantitative comparison (MAE in degrees) on the DiLiGenT-EV semi-real dataset and CW real dataset. Lower is better. The bolded values are the best performing values.}
    \label{tab:EventPS_dataset_result}
    \renewcommand{\arraystretch}{1.2}
    \resizebox{\textwidth}{!}{%
    \begin{tabular}{lcccccccccccccc}
    \toprule
    \multirow{2}{*}{Method} & \multicolumn{10}{c}{DiLiGenT-EV semi-real dataset} & \multicolumn{3}{c}{CW real dataset} & \multirow{2}{*}{Average} \\
    \cmidrule{2-10} \cmidrule{12-14}
    & Ball & Buddha & Cat & Cow & Goblet & Harvest & Pot1 & Pot2 & Reading & & Ball & Bunny & Tiger & \\
    \midrule
    EventPS-OP   & \textbf{4.12} & 16.65 & 8.53 & 25.31 & 15.52 & 35.11 & 9.64 & 13.32 & 22.49 & & 11.66 & 19.71 & 20.84 & 16.91 \\
    EventPS-FCN  & 7.37 & 15.84 & 8.45 & 17.41 & 12.79 & 26.09 & 10.84 & 13.87 & 20.75 & & 8.82  & 17.58 & 18.71 & 14.88 \\
    EventPS-CNN  & 4.53 & \textbf{12.06} & \textbf{6.44} & 16.55 & 9.91 & 24.97 & \textbf{7.69} & 10.33 & 18.18 & & 7.94  & \textbf{15.94} & 18.12 & 12.72 \\
    Ours & 4.52 & 12.45 & 6.69 & \textbf{13.58} & \textbf{9.83} & \textbf{23.63} & 7.95 & \textbf{9.89} & \textbf{17.21} & & \textbf{7.52}  & 16.10 & \textbf{17.94} & \textbf{12.24} \\
    \bottomrule
    \end{tabular}%
    }
\end{table*}

In EventPS, a null-space vector is first computed based on the light direction and the contrast threshold of the event camera, and this vector is then used as the input to the surface normal estimation algorithms. Given a known light trajectory, the light direction corresponding to each event timestamp can be determined, enabling the computation of the null-space vector regardless of the specific type of light trajectory. In contrast, our method assumes a fixed light trajectory during both training and inference. As a result, the light trajectory must remain consistent across training and validation. To accommodate this constraint, we prepared separate training and validation datasets for each light trajectory. For each configuration, the model was trained independently using the same set of objects with identical poses to ensure fair comparison.
\section{Experiments}\label{sec:experiments}

\begin{figure}[t]
    \centering
    \includegraphics[width=\linewidth]{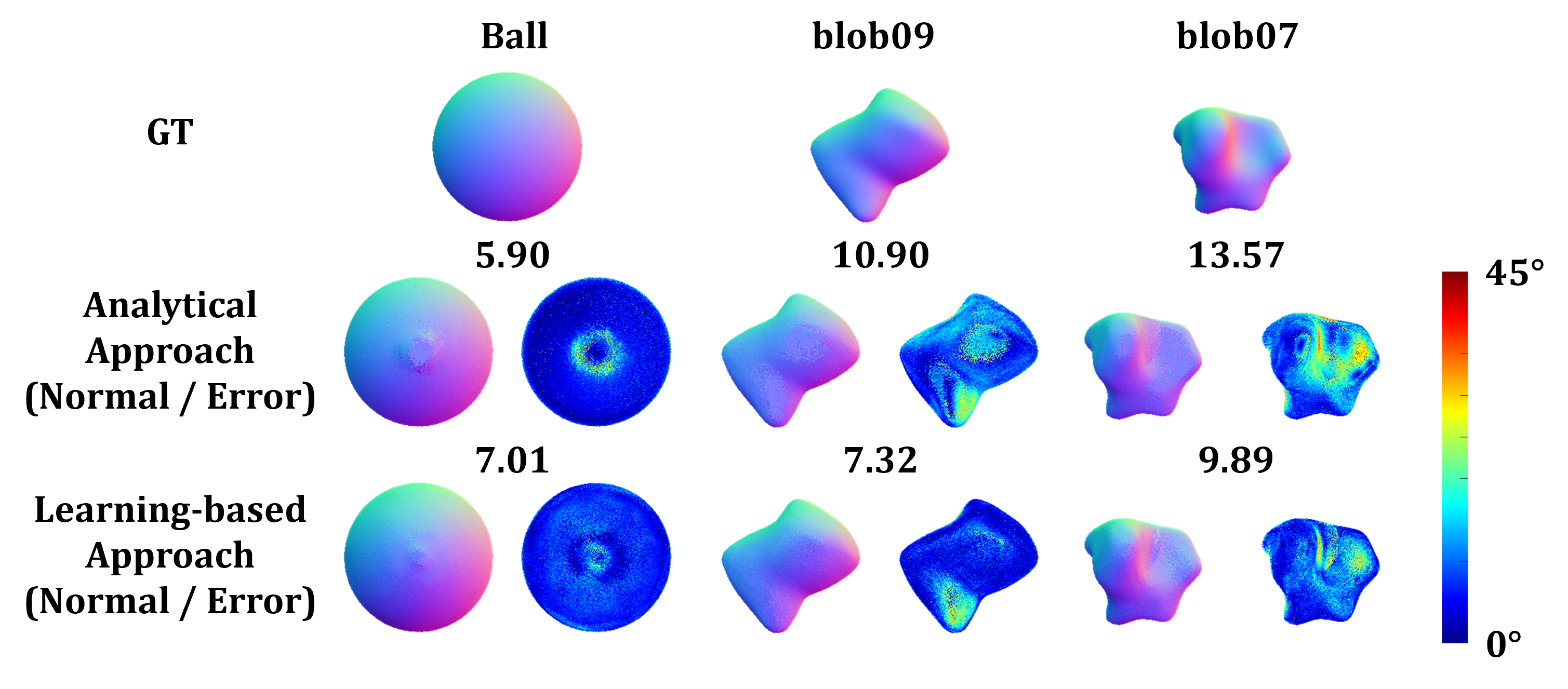}
    \caption{Comparison between analytical approach and learning-based approach. The number between predicted normal map and error map represents MAE.}
    \label{fig:analytical_vs_learning}
\end{figure}

\begin{figure}[t]
    \centering
    \includegraphics[width=\linewidth]{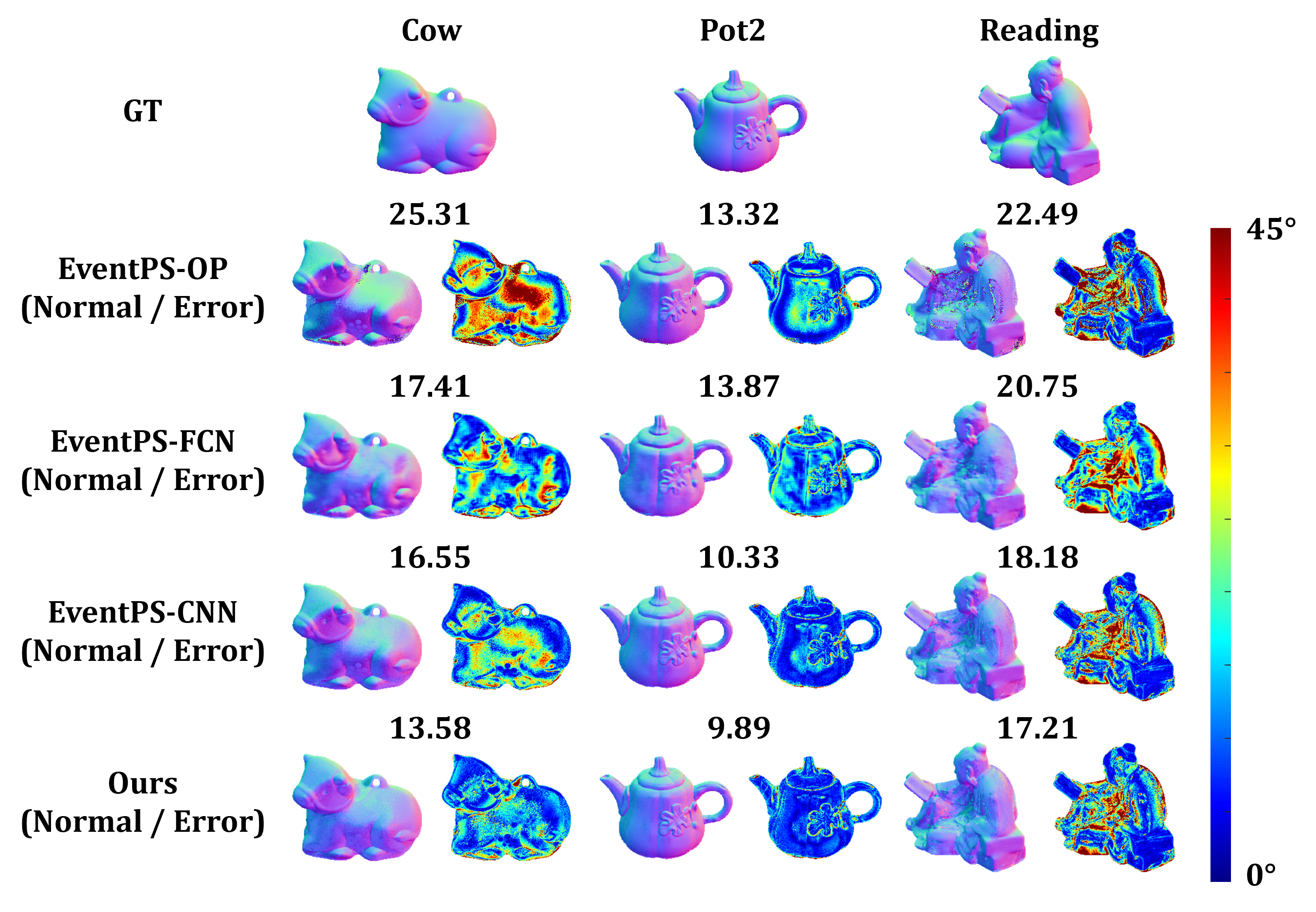}
    \caption{Evaluation results on DiLiGenT-EV semi-real dataset. The number between predicted normal map and error map represents MAE.}
    \label{fig:diligent_result}
\end{figure}

\begin{figure}[t]
    \centering
    \includegraphics[width=\linewidth]{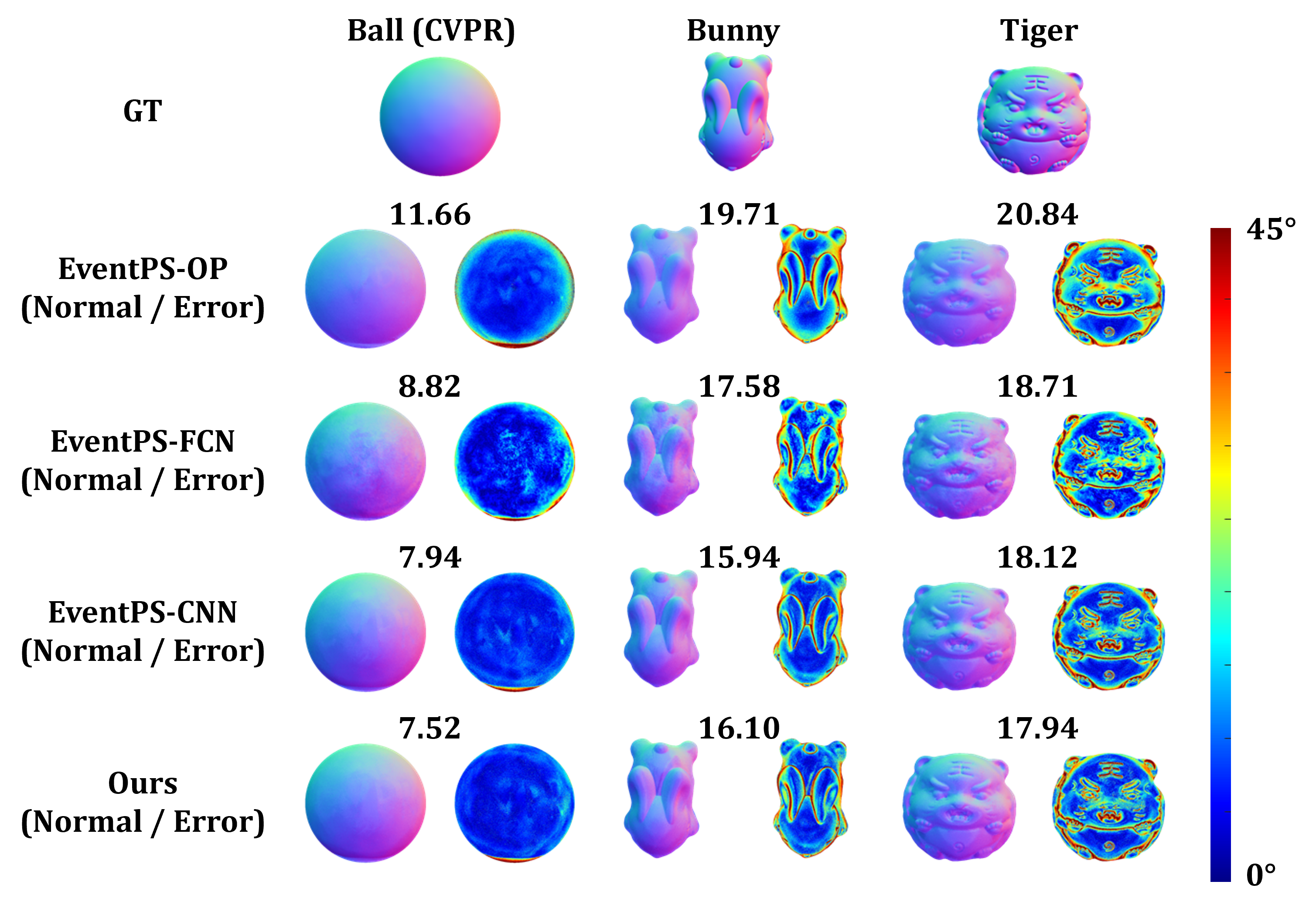}
    \caption{Evaluation results on CW real dataset. The number between predicted normal map and error map represents MAE.}
    \label{fig:cw_result}
\end{figure}

\subsection{Implementation details}
The DAVIS 346 event camera provides a maximum resolution of 346$\times$260 pixels. In our experiments, we applied center cropping to obtain 256$\times$256 input dimensions. The rotation period of the light source was set to approximately 1 second, and the average contrast threshold of the event camera was measured to be around 0.2. During training, the initial learning rate was set to 0.001, and the batch size was set to 256. To ensure fair comparison with our method, we implemented the three EventPS baselines: EventPS-OP, EventPS-FCN, and EventPS-CNN in Python, enabling them to be trained and evaluated on the same training and validation datasets as our method. 
For EventPS-CNN, we set the resolution of the observation map to 32$\times$32, with a batch size of 256. For EventPS-FCN, the number of time bins was set to 16, and the batch size was 1. All learning-based methods, including our method and the EventPS baselines, were implemented with PyTorch \citep{paszke2019pytorch}, and trained with NVIDIA GeForce RTX 4080 SUPER GPU for 250 epochs with the Adam optimizer \citep{kingma2014adam}.

\subsection{Evaluation on test dataset}

\paragraph{Evaluation metric}
Photometric stereo tasks are conventionally evaluated using the Mean Angular Error (MAE), which quantifies the average angular deviation between predicted surface normals and ground truth normals. Accordingly, we adopted MAE as our evaluation metric, where each angular error is computed as the angle between the predicted and ground truth normal vectors at each pixel coordinate.

\begin{table*}[tp]
    \centering
    \caption{Quantitative comparison (MAE in degrees) on the CCW real dataset. Lower is better. The bolded values are the best performing values.}
    \label{tab:MAE_ccw}
    \renewcommand{\arraystretch}{1.2}
    \resizebox{0.9\textwidth}{!}{%
    \begin{tabular}{lccccccccccccc}
    \toprule
    Method & Ball & blob01 & blob02 & blob03 & blob04 & blob06 & blob07 & blob08 & blob09 & blob10 & Average \\
    \midrule
    EventPS-OP   & \textbf{5.75} & 13.37 & 14.07 & 17.29 & 14.28 & 14.84 & 12.45 & 17.43 & 10.30 & 13.41 & 13.32 \\
    EventPS-FCN  & 8.30 & 9.57 & \textbf{10.11} & \textbf{11.19} & 10.41 & \textbf{9.76} & 10.46 & \textbf{11.44} & 8.33 & 9.30 & 9.89 \\
    EventPS-CNN  & 8.43 & 11.60 & 11.41 & 12.94 & 10.96 & 11.27 & 10.85 & 13.95 & 8.60 & 10.71 & 11.07 \\
    Ours         & 7.01 & \textbf{9.48} & 10.14 & 11.36 & \textbf{9.94} & 10.45 & \textbf{9.89} & 12.77 & \textbf{7.32} & \textbf{9.29} & \textbf{9.77} \\
    \bottomrule
    \end{tabular}%
    }
\end{table*}

\paragraph{Comparison with analytical approach}
As a preliminary evaluation, we validated the effectiveness of the learning-based approach, by comparing it with the analytical method based on cosine regression introduced in Section~\ref{sec:problem_formulation}. The evaluation was conducted using the CCW real dataset acquired through our data acquisition system. Fig.~\ref{fig:analytical_vs_learning} shows the predicted normal maps and the corresponding angular error maps for each approach. The results indicate that the learning-based approach achieves higher accuracy in surface normal estimation. In particular, it demonstrates enhanced robustness in regions affected by complex photometric effects such as interreflections, cast shadows, and specular highlights.

\begin{figure}[t]
    \centering
    \includegraphics[width=0.9\linewidth]{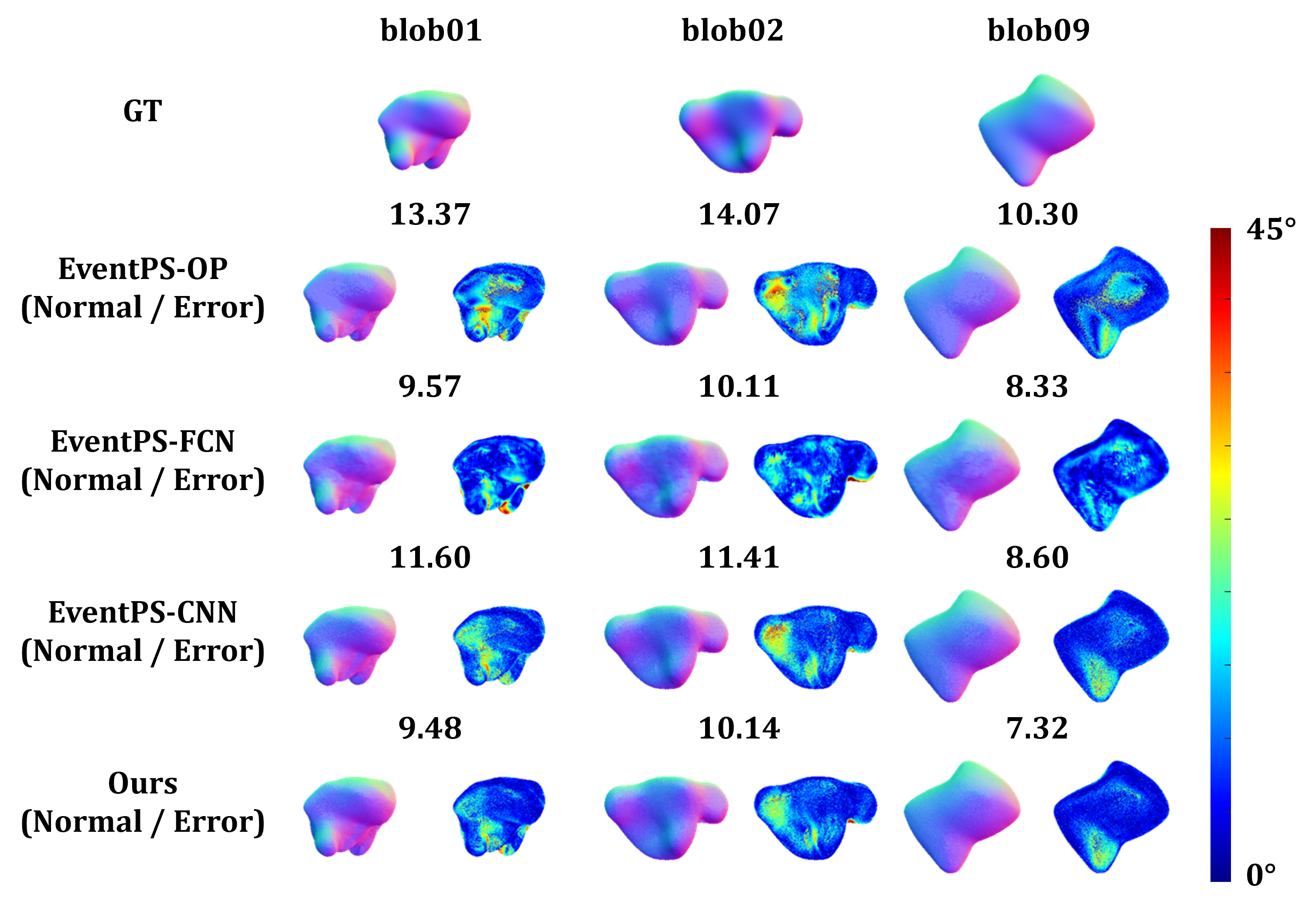}
    \caption{Evaluation results on CCW real dataset. The number between predicted normal map and error map represents MAE.}
    \label{fig:ccw_result}
\end{figure}

\paragraph{Evaluation on EventPS dataset}
Next, we conducted a quantitative comparison of the proposed method with the EventPS baselines. The evaluation was conducted on two test datasets provided by EventPS: the DiLiGenT-EV semi-real dataset, and the CW real dataset. For each object in the datasets, the average MAE between the predicted and ground truth surface normals was computed.
The MAE for each object is shown in Table~\ref{tab:EventPS_dataset_result}. On average, our method achieved the best performance with MAEs of 12.24. In particular, the proposed method demonstrated strong performance on objects exhibiting strong specularities due to shiny BRDFs (e.g., Cow, Pot2, Reading). For qualitative evaluation, we show three object examples of DiLiGenT-EV semi-real dataset in Fig.~\ref{fig:diligent_result}, and CW real dataset in Fig.~\ref{fig:cw_result}. The overall results of DiLiGenT-EV semi-real dataset are provided in Fig.~\ref{fig:all_results_diligent}.

\begin{figure}[t]
    \centering
    \includegraphics[width=0.8\linewidth]{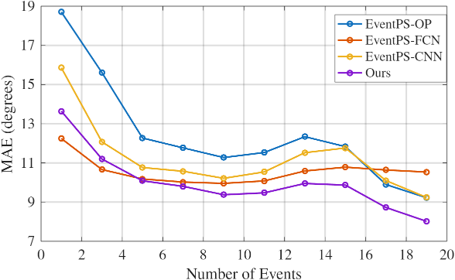}
    \caption{Quantitative evaluation showing the influence of event count on angular error for each method.}
    \label{fig:flat_area_a}
\end{figure}

\begin{figure}[t]
    \centering
    \includegraphics[width=0.85\linewidth]{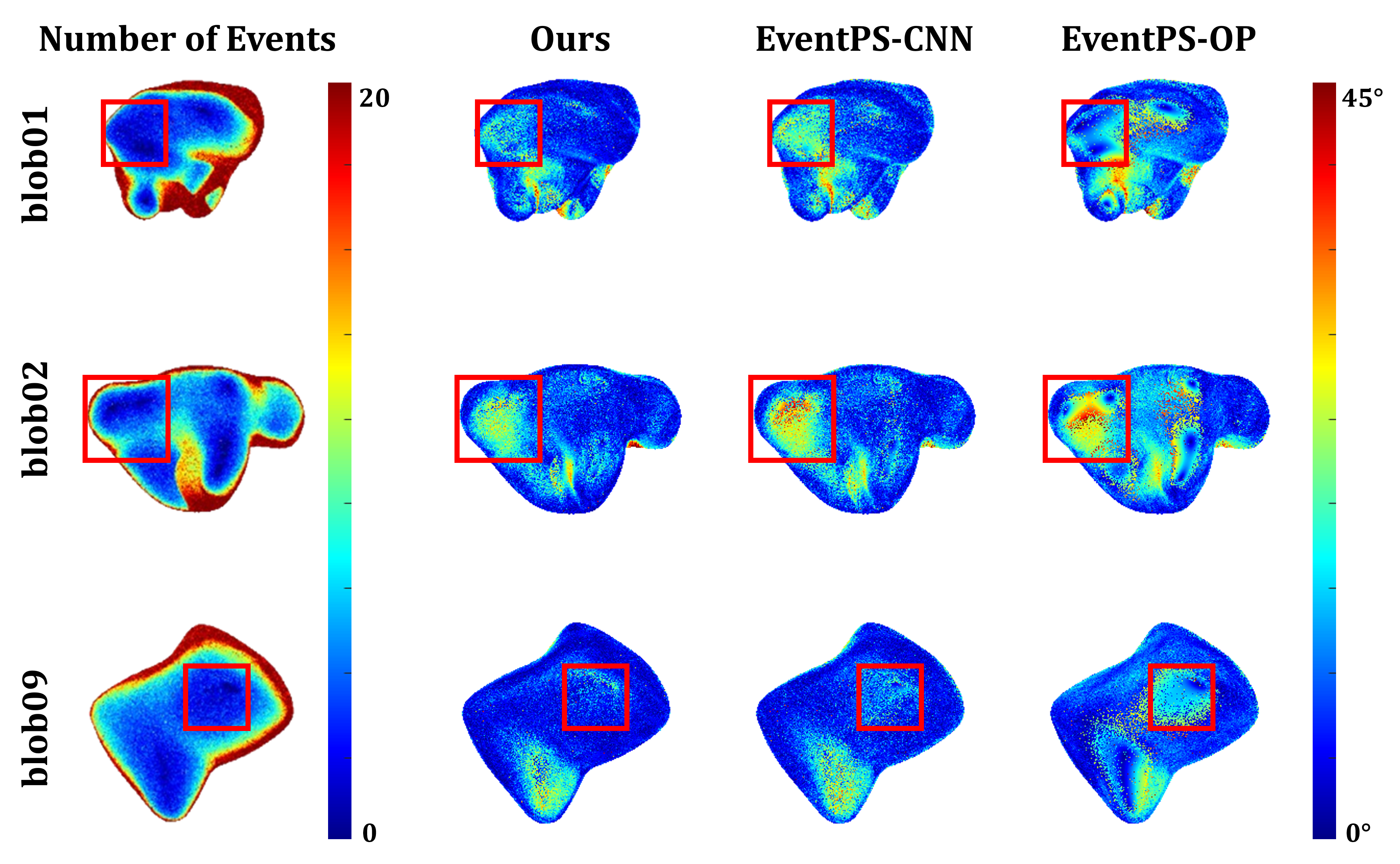}
    \caption{Qualitative comparison of error maps across methods, where the red boxes indicate regions with sparse event activity. Our method achieves the lowest angular error in these areas, demonstrating its robustness under low event density conditions.}
    \label{fig:flat_area_b}
\end{figure}

\paragraph{Evaluation on our data acquisition system}

To evaluate the general applicability of the proposed method, we conducted an evaluation on the CCW real dataset acquired using our data acquisition system. The MAE for each object is shown in Table~\ref{tab:MAE_ccw}. On average, our method achieved the best performance with MAEs of 12.24. Three object examples are shown in Fig.~\ref{fig:ccw_result}. From the error maps in Fig.~\ref{fig:ccw_result}, we observed that relatively high angular errors occur not only at object edges, but also in regions where events are sparsely generated. This suggests that low event density, often found in areas where surface normals align with the viewing direction, contributes to error. To investigate this phenomenon in a more structured manner, we grouped pixels based on the number of events accumulated at each location. Specifically, pixels were binned into intervals such as 1–2, 3–4, ..., 19–20 events, and the MAE was computed within each group. Fig.~\ref{fig:flat_area_a} presents the MAE computed for each event-count group across all methods, enabling a comparative analysis of performance under varying levels of event density. Fig.~\ref{fig:flat_area_b} shows the event count maps and corresponding error maps. Among the per-pixel methods including EventPS-OP, EventPS-CNN, and our proposed method, our proposed method consistently achieved the lowest MAE in regions with limited event activity, highlighting its robustness in low-signal scenarios. The overall results of CCW real dataset are provided in Fig.~\ref{fig:all_results_ccw}.

\begin{figure}[t]
    \centering
    \includegraphics[width=\linewidth]{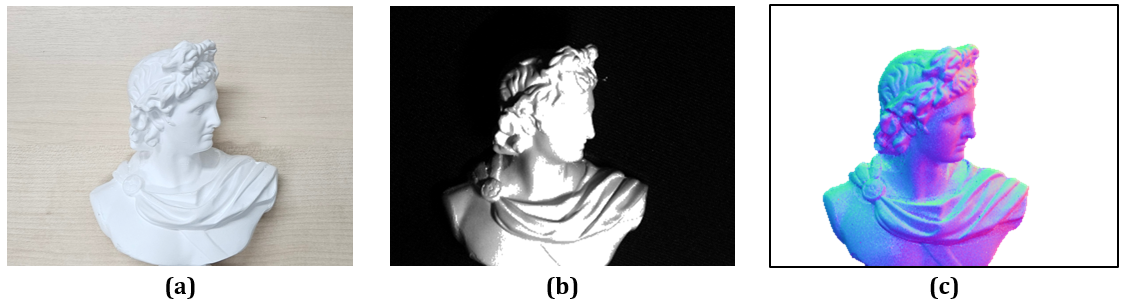}
    \caption{Surface normal estimation of a complex shape object under a high exposure scenario. (a) Captured image of the object, (b) Captured image under high exposure scenario. (c) Predicted normal map using event signal.}
    \label{fig:saturation}
\end{figure}

\paragraph{Evaluation under high dynamic range condition}
To evaluate the effectiveness of the event camera in high dynamic range scenarios, we conducted an experiment under conditions involving strong illumination and high exposure. Fig.~\ref{fig:saturation} shows the results of the estimated normal map of the object obtained using the event signal. Under such high exposure condition, conventional camera exhibited severe saturation, particularly in highly illuminated regions. This saturation led to a significant loss of photometric detail, which in turn hindered the accurate estimation of surface normals. In contrast, the event camera was able to capture meaningful signal changes even in extremely bright areas. As a result, the predicted normal map using the event data successfully preserved surface geometry, demonstrating the robustness of event-based methods in high dynamic range condition.



\section{Conclusion}\label{sec:conclusion}

In this work, we proposed a learning-based photometric stereo framework that estimates surface normals from event signals triggered by a single light source moving continuously around the event camera. Unlike conventional frame-based photometric stereo methods that rely on multiple discrete light sources and synchronized image captures, our approach demonstrates that dense surface normals can be acquired by observing the temporal variation in event streams caused by a single rotating illumination. This enables surface normal estimation using significantly simpler hardware setups. We formulated an analytical model for event-based photometric stereo and extended it to a learning-based per-pixel framework using a multi-layer neural network. We validated our method on the DiLiGenT-EV semi-real dataset, the CW real dataset, and our own CCW real dataset, achieving consistently superior performance compared to existing baselines, particularly in regions with sparse event activity or strong specularities. Additionally, we qualitatively demonstrated that our method remains robust under high dynamic range conditions.

\section*{ACKNOWLEDGMENTS} 
This research was supported by Culture, Sports and Tourism R\&D Program through the Korea Creative Content Agency grant funded by the Ministry of Culture, Sports and Tourism in 2024 (Project Name: Global Talent for Generative AI Copyright Infringement and Copyright Theft, Project Number: RS-2024-00398413, Contribution Rate: 90\%), and the National Research Foundation of Korea(NRF) grant funded by the Korea government(MSIT) (RS-2025-16072782, Contribution Rate: 10\%).


\clearpage
\appendix
\onecolumn
\section{All evaluation results}

\begin{figure}[h]
    \centering
    \includegraphics[width=0.95\textwidth]{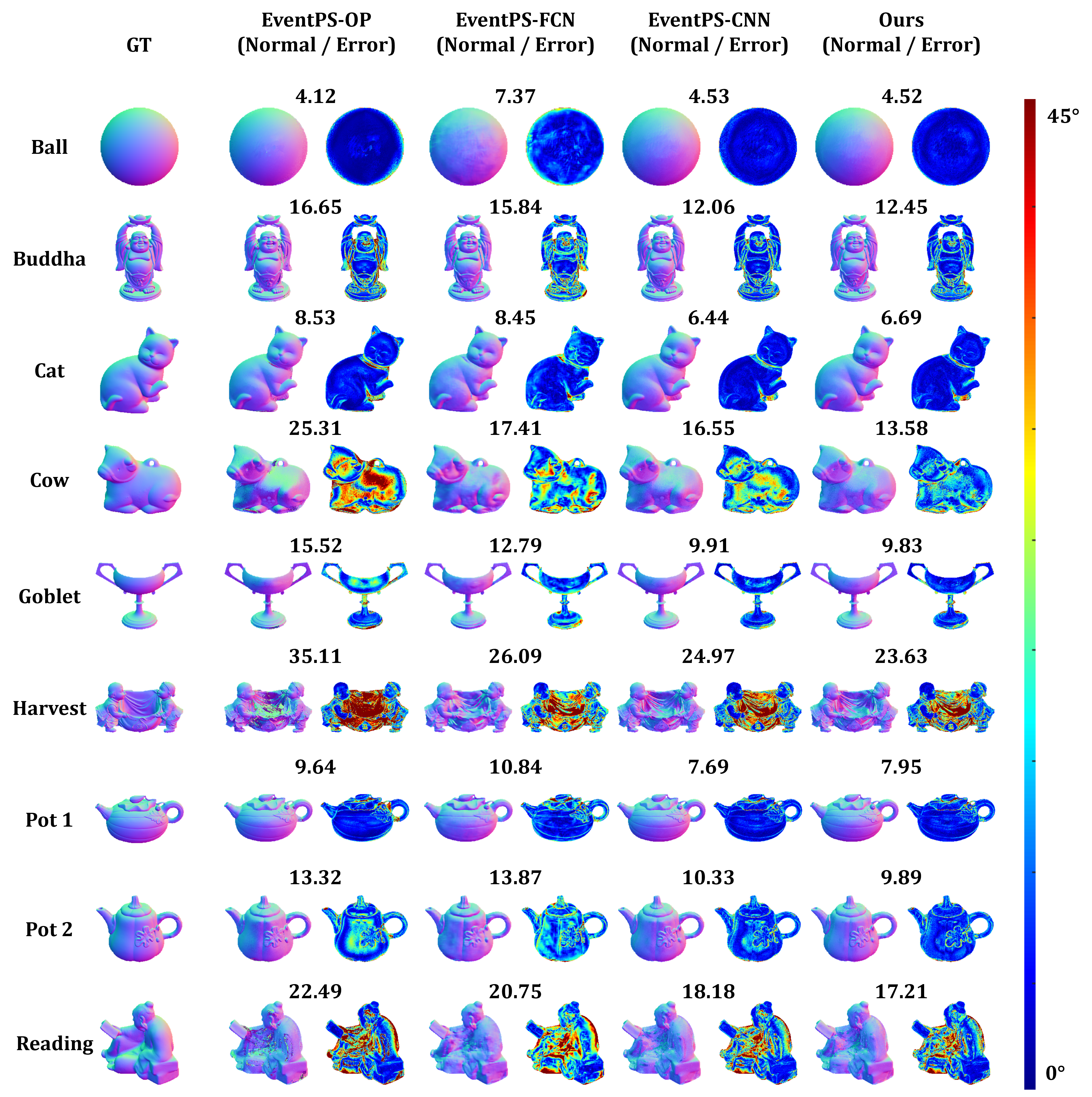}
    \caption{All evaluation results on DiLiGenT-EV semi-real dataset. The number between predicted normal map and error map represents MAE.}
    \label{fig:all_results_diligent}
\end{figure}

\clearpage
\begin{figure}[t]
    \centering
    \includegraphics[width=0.95\textwidth]{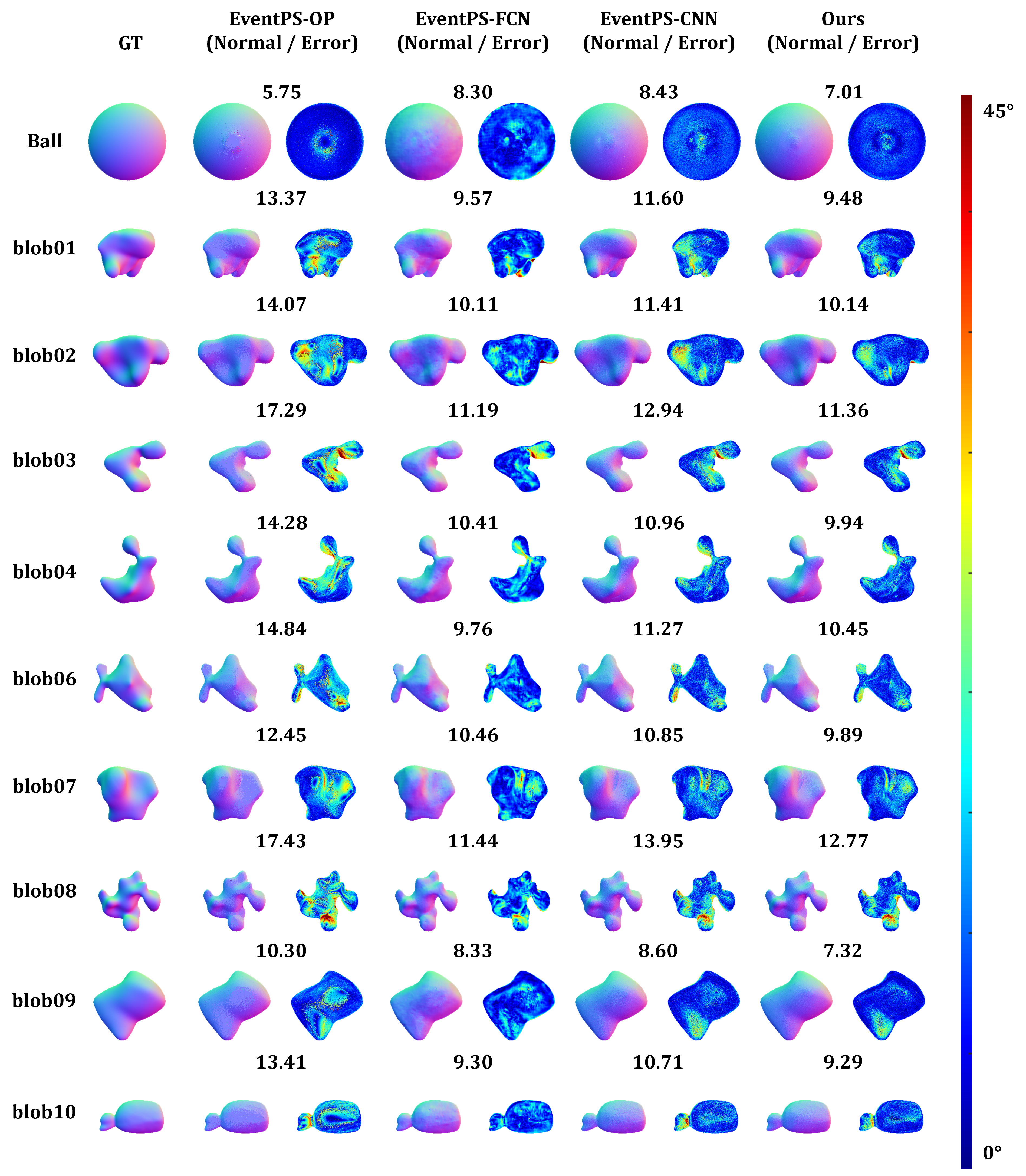}
    \caption{All evaluation results on CCW real dataset. The number between predicted normal map and error map represents MAE.}
    \label{fig:all_results_ccw}
\end{figure}

\twocolumn

\bibliographystyle{elsarticle-harv} 
\bibliography{main}

\end{document}